\documentclass[10pt,twocolumn,letterpaper]{article}

\usepackage{cvpr}              

\usepackage{times}

\usepackage{epsfig}
\usepackage{graphicx}
\usepackage{amsmath}
\usepackage{amssymb}
\usepackage{adjustbox}
\usepackage[utf8]{inputenc} 
\usepackage[T1]{fontenc}    
\usepackage{url}            
\usepackage{booktabs}       
\usepackage{amsfonts}       
\usepackage{nicefrac}       
\usepackage{microtype}      
\usepackage{xcolor}         
\usepackage{multirow}
\usepackage{tabu}
\usepackage{tabularx}
\usepackage{graphicx}
\usepackage{adjustbox}
\usepackage{amsopn}
\usepackage{capt-of}
\usepackage{todonotes}
\usepackage{subcaption}
\usepackage{arydshln}
\usepackage{bm}

\usepackage{pifont}
\newcommand{\cmark}{\ding{51}}%
\newcommand{\xmark}{\ding{55}}

\definecolor{linkc}{rgb}{0, 0.44, 0.74}
\definecolor{eqc}{rgb}{1, 0, 0}
\usepackage[pagebackref=false,breaklinks=true,letterpaper=true,colorlinks,urlcolor=linkc,citecolor=linkc,linkcolor=eqc,bookmarks=false]{hyperref}

\usepackage[capitalize]{cleveref}
\crefname{section}{Sec.}{Secs.}
\Crefname{section}{Section}{Sections}
\Crefname{table}{Table}{Tables}
\crefname{table}{Tab.}{Tabs.}

\def\cvprPaperID{*****} 
\def\confName{CVPR}
\def\confYear{2023}

\begin{document}

\title{Speed Co-Augmentation for Unsupervised Audio-Visual Pre-training}

\author{Jiangliu Wang\\
The Chinese University of Hong Kong\\
\and
Jianbo Jiao\\
University of Birmingham\\
\and
Yibing Song\\
Fudan University\\
\and
Stephen James\\
Dyson Robot Learning Lab\\
\and
Zhan Tong\\
Tencent AI Lab\\
\and
Chongjian Ge\\
The University of Hong Kong\\
\and
Pieter Abbeel\\
UC Berkeley\\
\and
Yun-Hui Liu\\
The Chinese University of Hong Kong
}
\maketitle

\begin{abstract}
This work aims to improve unsupervised audio-visual pre-training. 
Inspired by the efficacy of data augmentation in visual contrastive learning, we propose a novel speed co-augmentation method that randomly changes the playback speeds of both audio and video data.
Despite its simplicity, the speed co-augmentation method possesses two compelling attributes: (1) it increases the diversity of audio-visual pairs and doubles the size of negative pairs, resulting in a significant enhancement in the learned representations, and (2) it changes the strict correlation between audio-visual pairs but introduces a partial relationship between the augmented pairs, which is modeled by our proposed SoftInfoNCE loss\footnote{Our study~\cite{chongjian2023soft} also validates the effectiveness of the proposed ``SoftInfoNCE loss'' in single-modality contrastive learning.} to further boost the performance. 
Experimental results show that the proposed method significantly improves the learned representations when compared to vanilla audio-visual contrastive learning.
\end{abstract}

\section{Introduction}

Audio-visual contrastive learning~\cite{morgado2020audio,patrick-iccv21-compositions} for unsupervised pre-training has received growing attention  due to the observation that video content is usually accompanied by audio signals. The alignment of signals across audio and video forms a natural correspondence to benefit contrastive learning.
Under this framework, quite a few existing approaches~\cite{morgado2021robust,patrick-iccv21-compositions} focus on achieving better discrimination of positive and negative pairs to improve audio-visual representation learning. 
While promising results have been achieved, most works~\cite{ma2020active,morgado2021robust} apply data augmentations to each modality individually, which may potentially limit the diversity of the generated data views and restrict the potential of augmentation for contrastive learning.

\begin{figure}[t]
    \centering
    \includegraphics[width=1\linewidth]{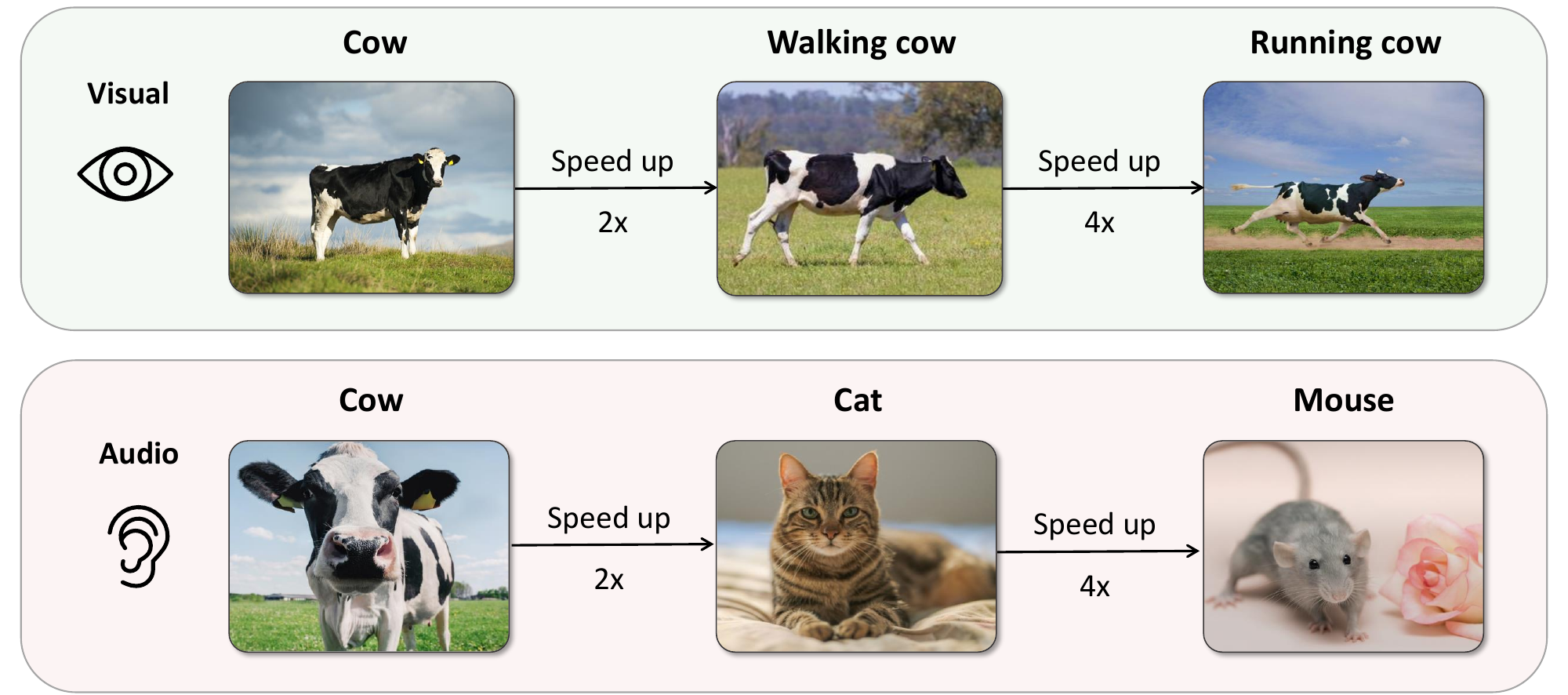}
    \caption{An intuitive example of \textit{semantic shift} after applying speed co-augmentation on audio-visual pairs. 
    Top: when speeding up video data, the semantic meaning of the content doesn't change drastically.
    Bottom: when speeding up audio data, the semantic meaning of the content changed drastically. 
    }
    \label{fig:tease}
\end{figure}

In this work, we propose a novel technique termed "speed co-augmentation" for unsupervised audio-visual pre-training, which involves modifying the playback speeds of both audio and visual data simultaneously. The speed co-augmentation method enhances the diversity of audio-visual pairs and doubles the number of negative pairs during training, which has been shown to be a crucial aspect of contrastive learning~\cite{chen-icml20-simclr}. Our experimental results demonstrate that this simple co-augmentation method yields a significant performance improvement of 10.0\% over the baseline audio-visual contrastive learning approach on the HMDB51~\cite{kuehne2011hmdb} dataset.

Meanwhile, it was observed that after the speed co-augmentation, the audio and video pairs derived from the same clip are no longer strictly positively related, as is commonly assumed.
As an intuitive special example (Fig.~\ref{fig:tease}), a sped-up \textit{cow} still visually looks like a \textit{cow}, but a sped-up \textit{cow} may auditorily sound like a \textit{cat}.
To generalize, we posit that there exists a partial relationship between the augmented audio-visual pairs, which is influenced by the degree of speed augmentation applied.
To capture this relationship, we introduce a cross-affinity module that automatically learns the audio-visual correlations across different views. The resulting learned correlations quantitatively measure the audio-visual consistency and are employed for computations of SoftInfoNCE loss, leading to a further performance boost. Combining the proposed speed augmentation and the cross-affinity module, we present a Speed-augmented visual-audio Contrastive Learning framework, which we call \emph{SvaCLR}.


\section{Method}


Our target is to train video and audio encoders via unsupervised contrastive learning.
Given an aligned pair $(v, a)$, we apply speed-up augmentations on both audio $a$ and video $v$ data to synthesize two additional views (\ie, $\widetilde{v}$ and $\widetilde{a}$). These audio and video samples are then fed into the audio and video encoders $f(\cdot)$ and $g(\cdot)$ to extract representations $y$. We then project the video and audio representations separately via projectors $h_v(\cdot)$ and $h_a(\cdot)$. The projected embeddings $z$ are then utilized to compute the contrastive InfoNCE loss~\cite{oord2018representation}. In parallel, we introduce a cross-affinity module to model the audio-visual embedding correlations. The modeled correlations are used to reweigh the InfoNCE loss, \ie, the proposed SoftInfoNCE loss,  when learning audio-visual representations. In the following, we first introduce the speed-up augmentation with the vanilla InfoNCE loss and then introduce the cross-affinity module to re-weigh the InfoNCE loss (\ie, SoftInfoNCE loss).

\subsection{Speed co-augmentation}\label{sec:speed}

For speed co-augmentation, we use a speed library to diversify training data pairs.
We use $\mathcal{T}$ to represent the speed co-augmentation set in which the maximum speed is denoted by $S$. Each time, two speed augmentation factors for the audio and video data are selected randomly from $\mathcal{T}$ and are applied to each data, respectively. In practice, the proposed speed co-augmentation is implemented by applying different sampling rates of the audio and video samples.

Before computing the contrastive InfoNCE loss~\cite{oord2018representation}, we project the video and audio representations separately via projectors. We use one video projector $h_v(\cdot)$ to connect the video encoder and use one audio projector $h_a(\cdot)$ to connect the audio encoder. The projected representations are then utilized to compute the contrastive InfoNCE loss as follows:

\begin{equation}\label{eq:nce}
    L(i,j)=\frac{\exp({z_i\cdot z_j}~/~{\eta})}{\exp({z_i\cdot z_j}~/~{\eta}) + \sum\limits_{j=1 \atop j\neq i}^N \exp({z_i\cdot z_j}~/~{\eta})}
\end{equation}

where $z_i=h_a(y_i)$ is the audio projection,  $z_j=h_v(y_j)$ is the video projection, and $\eta$ is a constant temperature value. The dot product measures the similarity between the projected audio and video representations. For the input audio $a_i$, the summation term is computed by utilizing all the video clips $v_j$, as long as $a_i$ and $v_j$ are from different samples (\ie, unpaired).

\begin{figure}[t]
    \centering
    \includegraphics[width=\linewidth]{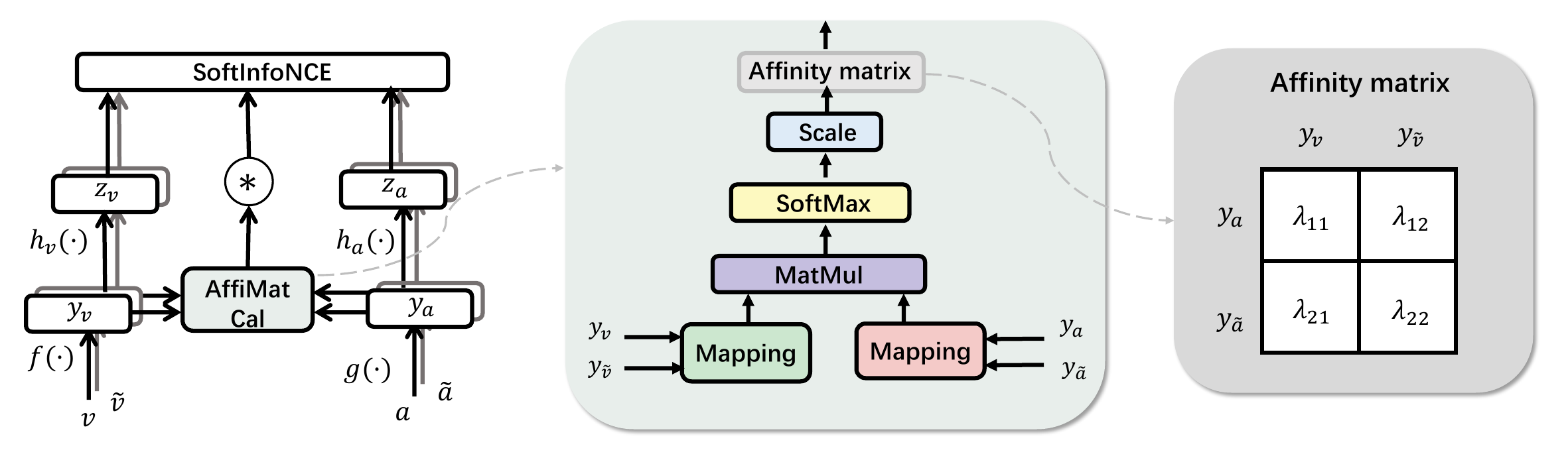}
    \vspace{-2mm}
    \caption{The proposed cross-affinity module for SoftInfoNCE loss computation. The cross-modality attention module takes video and audio representations as input where there are co-augmented audio-visual data. The output is a cross-modality affinity matrix shown on the right. Each element in this matrix represents the correlations between audio and video for each input signal.}
    \label{fig:attmatrix}
\end{figure}

\subsection{Cross-affinity module} \label{sec:soft}

We propose a cross-affinity module to measure the correlations between the augmented video and audio representations. Fig.~\ref{fig:attmatrix} illustrates the proposed module. Given the audio embedding $y_i$ and the video embedding $y_j$, the cross-modality attention $\lambda(a_i^{\mathcal{\tau}_1},v_j^{\mathcal{\tau}_2})$ can be computed as follows:
\begin{equation}\label{eq:cross}
    \lambda(a_i^{\mathcal{\tau}_1},v_j^{\mathcal{\tau}_2})=\operatorname{softmax}\left[{l(y_i)\times l(y_j^\textrm{T})}\right],
\end{equation}
where $l(\cdot)$ is a mapping with learnable parameters. The projected video and audio embeddings are correlated via the matrix multiplication operation. In practice, three different mapping function are examined, including identity mapping, linear mapping and nonlinear mapping, from those we find the identity mapping achieves the best results.  
We speculate this is because heavier mapping could deteriorate the ability of encoders to learn general representations.

We compute the cross-modality attention in Eq.~\ref{eq:cross} for one co-augmented audio-visual view. The cross-modality affinity can be formulated as a two-by-two matrix (as shown in Fig.~\ref{fig:attmatrix} right). Each element in this matrix represents the correlation between the speed-augmented audio and video views. By using these elements, we reweigh the contributions of each co-augmented audio-visual view when computing the contrastive loss.

\subsection{Training with SoftInfoNCE} \label{sec:network}

Following~\cite{patrick-iccv21-compositions,ma2020active}, we use a 9-layered 2D ResNet~\cite{he2016deep} as the audio encoder and R(2+1)D-18~\cite{r2plus1d_cvpr18} as the video encoder. The projector is a two-layered multilayer perceptron (MLP). The training process is end-to-end, without using a two-stage setting as in previous works~\cite{morgado2021robust, morgado2020audio}.

Given a batch of audio-visual pairs $\mathcal{A}$ and $\mathcal{V}$, where both $\mathcal{A}$ and $\mathcal{V}$ contain $N$ samples, we denote the speed-up augmentation set as $\mathcal{T}$, from which we can sample augmentations $\mathcal{\tau} \sim p(\mathcal{T})$. 
The encoders and projectors are trained with the following SoftInfoNCE loss:

\begin{equation}
\resizebox{0.9\hsize}{!}{
$ \mathcal{L}(f,g,\mathcal{A},\mathcal{V})=\mathbb{E}_{(\mathcal{\tau}_1,\mathcal{\tau}_2)\sim p(\mathcal{T})}\left[\frac{1}{N^2}\sum_{i=1}^N\sum_{j=1}^N \lambda(a_i^{\mathcal{\tau}_1},v_j^{\mathcal{\tau}_2})\cdot L(g(a_i^{\mathcal{\tau}_1}),f(v_j^{\mathcal{\tau}_2})) \right]$
}
\label{eq:soft}
\end{equation}
where $L(\cdot,\cdot)$ is the contrastive InfoNCE loss function as illustrated in Eq.~\ref{eq:nce}, $a_i \in \mathcal{A}$, and $v_j \in \mathcal{V}$. The cross-modality attention $\lambda(\cdot,\cdot)$ takes the audio and video signal as input and measures their correlations. The output correlation value further reweighs the contrastive loss during the training process consequently. 

\section{Experiments}
\subsection{Implementation details}  \label{sec:imple}

We use Kinetics-Sounds (K-Sounds)~\cite{arandjelovic2017look}, Kinetics-400 (K400)~\cite{kay2017kinetics}, and VGGSound~\cite{chen2020vggsound} as the pre-training datasets to evaluate the effectiveness of our proposed approach. 

For pre-training, we use Stochastic Gradient Descent (SGD) as the optimizer.
We use 10 epochs to warm up the learning rate from $64\times10^{-3}$ to $64\times10^{-2}$ and then using a cosine learning rate decay to $10\times10^{-2}$ in the remaining 90 epochs. 
Training is done on 64 V100 GPUs with a mini-bath size of 8 on each, resulting in total batch size of 512. The total training time is around 30 hours for 100 epochs. 

For fine-tuning, we follow the fine-tuning setting of GDT~\cite{patrick-iccv21-compositions}.
SGD is used as the optimizer and the initial learning rate is set to $2.5\times10^{-3}$, where it is warmed up to $2\times10^{-2}$ in the first two epochs, and decreased by $5\times10^{-2}$ at 6 and 10 epochs. Training is stopped at 12 epochs.

\begin{table}[t]
	\centering
	\caption{Effectiveness of the proposed speed co-augmentation and cross-affinity module. 
    Speed $s=[a,b]$ represents that the lower bound speed is $a$ while the upper bound is $b$. 
	}
	\label{tab:speed}
    \begin{adjustbox}{max width=\linewidth}
	\begin{tabular}{cccccc}
		\toprule
    \multicolumn{4}{c}{Pre-training Experimental Setup} &
				\multicolumn{2}{c}{Downstream Acc.} \\
				\cmidrule(lr){1-4}
				\cmidrule(lr){5-6}
		speed & Re-weight & Speed  & Loss &HMDB51 & K-sounds@1\\
		\midrule
		\xmark & \xmark & -  &  InfoNCE & 54.2 &  2.6\\
        \midrule
		\cmark & \xmark  & $s=[1,2]$ & InfoNCE & 63.8 (+9.6) &  3.1 (+0.5)\\
        \cmark & \xmark & \textbf{$s=[1,4]$}   &InfoNCE & \underline{65.1} (+10.9) & \underline{3.5} (+0.9) \\
        \cmark  & \xmark & $s=[1,6]$  & InfoNCE & 64.2 (+10.0) & 3.2 (+0.6)\\
        \midrule
        \cmark  & \cmark  & \bm{$s=[1,4]$} & \textbf{SoftInfoNCE} &\textbf{66.1 (+11.9)} & \textbf{4.6 (+2.0)} \\
		\bottomrule
	\end{tabular}
    \end{adjustbox}
\end{table}

\subsection{Ablation studies} \label{sec:ablation}

{ \bf Speed co-augmentation.} We investigate different augmentations for audio and video data from the same speed distribution in Table \ref{tab:speed}. It can be seen that: \textbf{(1)} compared to the no speed-up augmentation, using only two speed candidates can already significantly improve the action recognition performance from 54.2\% to 63.8\%. This validates the effectiveness of the proposed speed-up augmentation. \textbf{(2)} When audio-video pairs are sped up from the same distribution, best action recognition accuracy is achieved when $s=[1,4]$. Both $s=[1,2]$ and $s=[1,6]$ settings perform inferior than the $s=[1,4]$ setting. We suspect that this is because compared to $s=[1,2]$, using $s=[1,4]$ provides more view synthesis while $s=[1,6]$ could be a bit difficult for the network to learn useful semantic representations.

\begin{figure}[t]
    \centering
    \includegraphics[width=1\linewidth]{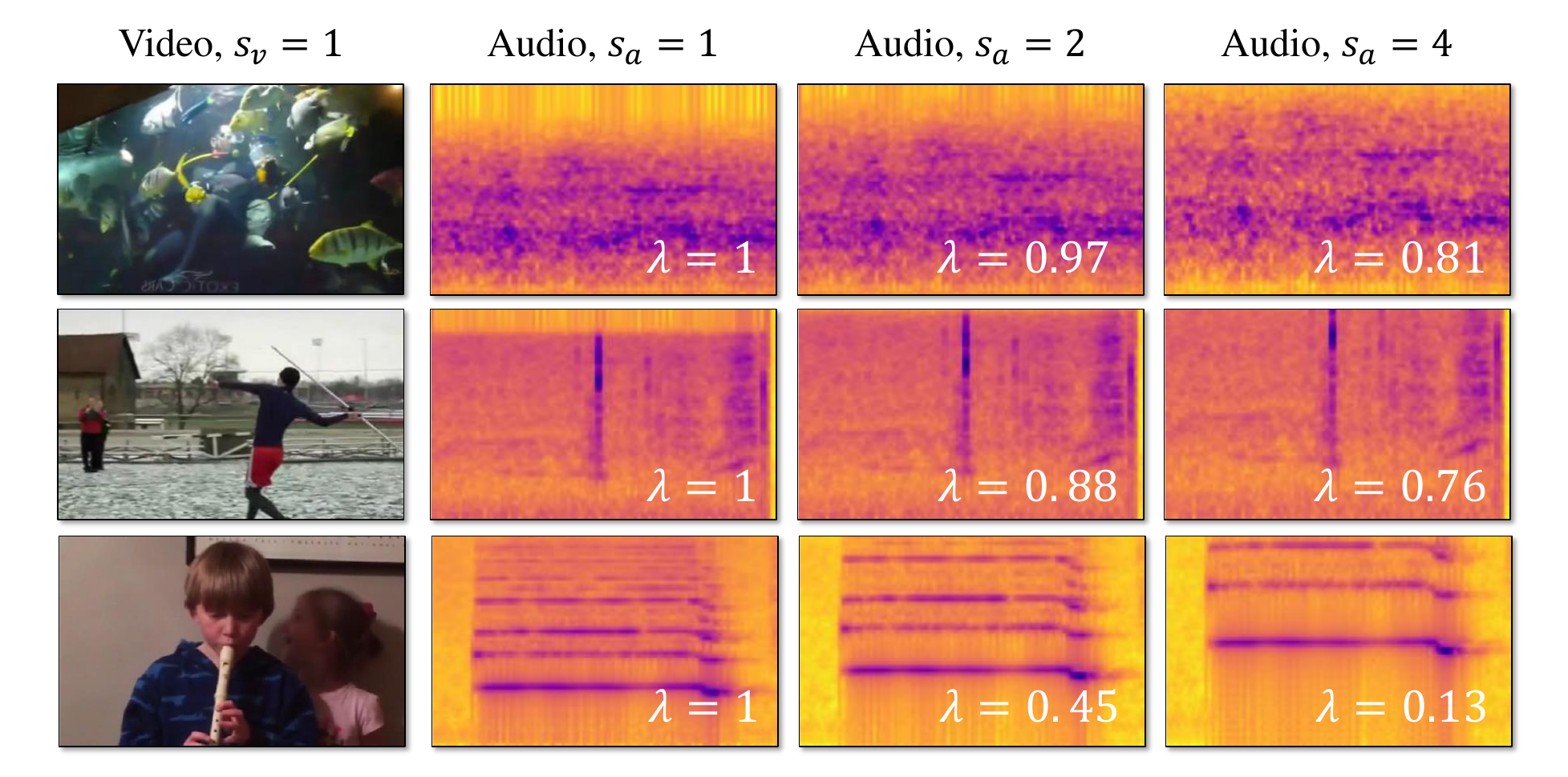}
    \caption{Examples of the visual-audio pairs with speed co-augmentation and their partial relationship weights. For each sample from left to right: one frame from the video clip, spectrogram of the corresponding audio, spectrogram of the augmented audio with speed 2, and spectrogram of the augmented audio with speed 4. $\lambda$ reflects the degree of correlations learned by our proposed cross-affinity module.
}
    \label{fig:spec_vs_weight}
\end{figure}

{ \bf Cross-affinity module.}
Based on the best speed augmentation setting $s_a=s_v=[1,4]$, we investigate the effectiveness of the proposed cross-affinity module in Table~\ref{tab:speed}. It can be seen that with the proposed cross-affinity module and the SoftInfoNCE loss design, the performances can be further improved significantly (\eg, +11.9\% for HMDB51). To better understand what the cross-affinity module learns, we visualize the visual-audio pairs with the speed augmentation and their corresponding relationship weights in Fig.~\ref{fig:spec_vs_weight}. It can be seen that videos with manifested audio signals are more sensitive to the proposed speed augmentation, where the weight between the video and augmented audio is relatively small after the speed transformation.                 

\begin{table}[htbp!]
\centering
	\caption{ Comparison with previous approaches on UCF101~\cite{soomro2012ucf101} and HMDB51~\cite{kuehne2011hmdb} datasets.} 
   \vspace{-5pt}
	\begin{center}
		\begin{adjustbox}{max width=\linewidth}
			\begin{tabular}{lcccccc}
				\toprule
				\multicolumn{1}{l}{\multirow{2}{*}{Method}} &
				\multicolumn{3}{c}{Pre-training Experimental Setup} &
				\multicolumn{2}{c}{Downstream Acc.} \\
				\cmidrule(lr){2-4}
				\cmidrule(lr){5-6}
				~ &  Architecture  & Dataset (duration) & Resolution & HMDB51 & UCF101  \\
				\midrule
                ACC~\cite{ma2020active} & R3D-18 & K-Sounds (2d)  & $224\times224$  & 40.6 & 77.2 \\
                \textbf{SvaCLR (Ours)} & R3D-18 & K-Sounds (2d)  & $128\times128$ & \textbf{54.9} & \textbf{86.4} \\
                \midrule
                $L^3$-Net~\cite{arandjelovic2017look} & VGG-16 & K400 (28d)  &  - & 40.2 & 72.3 \\
				PEMT~\cite{lee2021parameter} & SlowFast & K400 (28d) & $128\times128$  & - & 85.2 \\
    			GDT~\cite{patrick-iccv21-compositions}  & R(2+1)D-18 & K400 (28d)  & $128\times128$& 62.3 & 90.9 \\ 
                SvaCLR (Ours) & R(2+1)D-18 &  K400 (28d) & $128\times128$   & \underline{66.1} & \underline{91.5} \\
                 \textbf{SvaCLR (Ours)} & R(2+1)D-18 &  VGGSound (23d) & $128\times128$    & \textbf{67.2} &  \textbf{92.0} \\
                \midrule
				Multisensory~\cite{owens2018audio} & R3D-18 & K400 (28d)  & $224\times224$ & - & 82.1 \\
				SeLaVi~\cite{asano2020labelling} & R(2+1)D-18 & K400 (28d) & $224\times224$ & 47.1 & 84.2 \\
                XDC~\cite{alwassel2020self} & R(2+1)D-18 & K400 (28d) & $224\times224$  & 52.6 & 86.2 \\ 
				AVTS~\cite{korbar2018cooperative}  & MC3-18 & K400 (28d) & $224\times224$  & 56.9 & 85.8 \\
    			STiCA~\cite{patrick-iccv21-compositions} & R(2+1)D-18 & K400 (28d) & $224\times224$ & 60.5 & - \\
				AVID~\cite{morgado2020audio}  & R(2+1)D-18 & K400 (28d)  & $224\times224$ & 60.8 & 87.5 \\
				ACC~\cite{ma2020active} & R3D-18 & K400 (28d)  & $224\times224$ & 61.8 & 90.2 \\ 
                SvaCLR (Ours) & R(2+1)D-18 &  K400 (28d) & $224\times224$    & \underline{66.8} & \underline{92.2} \\
                \textbf{SvaCLR (Ours)} & R(2+1)D-18 &  VGGSound (23d) & $224\times224$  & \textbf{67.8} & \textbf{92.6} \\
				\bottomrule
			\end{tabular}
		\end{adjustbox}
	\end{center}
	\label{tab:main_sota}
\end{table}

\subsection{Comparison with previous works} \label{sec:sota}

{\flushleft \bf Action recognition.}
We compare our proposed SvaCLR with other approaches in Table~\ref{tab:main_sota}. 
From the results we can see that: \textbf{(1)} When pre-trained on a medium-scale dataset, K-Sounds, our approach significantly outperforms previous works. We improve performances on UCF101 and HMDB51 datasets by large margins, 9.2\% and 14.3\%. This demonstrates that our proposed co-augmentation method enlarges the diversity of the training views and benefits contrastive learning a lot. 
\textbf{(2)} 
Our approach demonstrates great scalability in terms of dataset size.
When pre-trained on a large dataset K400, our approach exceeds the state-of-the-art audio-visual representation learning approach GDT~\cite{patrick-iccv21-compositions} by a large margin, especially on the HMDB51 dataset, where we outperform GDT by 3.8\%.
Note that GDT applies hierarchical data augmentations while we only use one-speed augmentation. \
\textbf{(3)} Our approach also demonstrates scalability in terms of resolution. We can further improve the performances by using a large input size.

\paragraph{Audio-Video Retrieval.} To further evaluate the cross-modality ability of the proposed approach, we propose to use an audio-video retrieval task on K-Sounds~\cite{arandjelovic2017look}. 
We compare to audio-visual contrastive learning with vanilla InfoNCE loss and current state-of-the-art GDT~\cite{patrick-iccv21-compositions} in Table~\ref{tab:video-audio-retrieval}. We show that our approach achieves the best performances on both audio-to-video retrieval task and video-to-audio retrieval task. It is interesting to note that pre-trained on K-sounds can achieve better performance to retrieve top-1 nearest neighbor. But its ability to generalize to more visual-audio pairs is restricted that it performs worse than pre-training on a larger dataset K400 to retrieve top-5, top-10, and top-20 nearest neighbors.  

\begin{table}[h]
    \centering
    \caption{Comparison with other approaches on video-audio retrieval task.
    ``Baseline'' represents the vanilla audio-video contrastive learning with InfoNCE loss.}
    \vspace{-1.8mm}
     \begin{adjustbox}{max width=\linewidth}
    \begin{tabular}{llccccccccc}
         \toprule
         Method  & Pre-train Dataset &
         \multicolumn{4}{c}{Video $\rightarrow$ Audio} &
          \multicolumn{4}{c}{Audio $\rightarrow$ Video} \\
         \cmidrule(lr){3-6}
         \cmidrule(lr){7-10}
         ~ &  ~ & R@1 & R@5 & R@10 & R@20 & R@1 & R@5 & R@10 & R@20 \\
         \midrule
         Baseline &  K400 &2.6 & 13.8  &  26.1 & 44.3
         & 3.6 & 13.7 & 25.3 & 41.0\\
         GDT~\cite{patrick-iccv21-compositions} & K400 & 3.1 & 14.3 & 24.6 & 40.8 & 3.6 & 16.1 & 26.4 & 43.9\\
         \midrule
         \textbf{SvaCLR (Ours)} & K-Sounds & \textbf{5.1} & 14.1 & 25.4 & 42.3 &
         \textbf{4.7} & 16.0 & 27.2 & 43.6 \\
         \textbf{SvaCLR (Ours)} & K400 & 4.6 & \textbf{17.2} & \textbf{28.3} & \textbf{44.6}
         & 4.2 & \textbf{16.4} & \textbf{27.6} & \textbf{44.3} \\
         \bottomrule
    \end{tabular}
    \label{tab:video-audio-retrieval}
    \end{adjustbox}
\end{table}

\section{Conclusions} \label{sec:conclusion}
We proposed a speed co-augmentation method for unsupervised audio-visual pre-training. We observed that speed co-augmentation leads to a partial relationship between audio-visual pairs. 
To combat this, we propose a cross-affinity module, which can adaptively 
model the cross-modality partial relationship and further improve performances. Extensive experimental results show that our approach significantly improves the performances.

{\small
\bibliographystyle{ieee_fullname}
\bibliography{egbib}
}

\end{document}